\newcommand{\comment}[1]{}

\comment{

######################################################

Revisions:

-> Ensure format works with 1.5x spacing

REVIEWER #1 
-> Add comment on exploratory nature of work and that this is not proposed as ready for practice
-> Explain that offline RL can incorporate additional sensor information without being explicitly programmed to do so
-> reiterate the overall significance to JBI

REVIEWER #2
-> More prior knowledge on offline reinforcement learning is required
-> make statement around 1/10th of data more clear
-> More clearly explain that individual models were used for age group sectio
-> Would sampling rate affect performance? How to justify not doing an experiment on this? 
-> Add more context on what missing data is? 
-> Better to develop age group learning and modelling separately?
-> Add more of a comment on using retrospective data

REVIEWER #3 
-> Reference DQN algorithm on page 13
-> Add reference to code with revised implementation for SAC-RNN
-> Provide some reference for usable T1D datasets
-> Explain choice of no optimisation for offline RL
-> Add more explanation as to why performance is so different between age groups
-> Change last section to discussion and conclusions

REVIEWER #4
-> Add review of work using supervised learning for blood glucose control algorithms i.e. model based? 
-> Justify the statement "In moving towards use in hybrid closed loop systems, online RL performance is likely to deteriorate as the complexity of the training environment increases"
-> Justify the selection of the reward function and explain the alternative choices
-> Explain the following concepts in more detail:
    -> hybrid closed loop systems 
    -> reinforcement learning 
    -> offline and online RL 
-> explain the acronym MDP 
-> reword "In offline RL, the agent is incapable of environmental interaction and instead must rely on samples generated by a demonstrator, such as a PID algorithm, to build an understanding of the MDP"

#####################################################
}

\documentclass[preprint, 12pt]{elsarticle}

\usepackage{graphicx}
\graphicspath{{}}

\usepackage{booktabs}
\usepackage{array}

\usepackage{subcaption}

\usepackage{multicol}

\usepackage{amssymb, amsmath}
\DeclareMathOperator*{\argmax}{arg\,max}

\usepackage{hyperref}
\usepackage{color}

\usepackage{float}

\usepackage{amssymb}
\begin{document}

\begin{frontmatter}

%% Title, authors and addresses

%% use the tnoteref command within \title for footnotes;
%% use the tnotetext command for theassociated footnote;
%% use the fnref command within \author or \address for footnotes;
%% use the fntext command for theassociated footnote;
%% use the corref command within \author for corresponding author footnotes;
%% use the cortext command for theassociated footnote;
%% use the ead command for the email address,
%% and the form \ead[url] for the home page:
\title{Offline Reinforcement Learning for Safer Blood Glucose Control in People with Type 1 Diabetes}

\author{Harry Emerson\fnref{label1}\corref{cor1}}
\ead{harry.emerson@bristol.ac.uk}

\author{Matthew Guy\fnref{label2}\corref{cor2}}
\ead{matthew.guy@uhs.nhs.uk}

\author{Ryan McConville\fnref{label1}\corref{cor2}}
\ead{ryan.mcconville@bristol.ac.uk}

%% use optional labels to link authors explicitly to addresses:
\address[label1]{
    University of Bristol, 
    1 Cathedral Square, 
    Bristol,
    BS1 5TS, 
    United Kingdom
}

\address[label2]{
    University Hospital Southampton, 
    Tremona Road,
    Southampton,
    SO16 6YD,
    Hampshire, 
    United Kingdom
}

\cortext[cor1]{Corresponding author}

\begin{abstract}
The widespread adoption of effective hybrid closed loop systems would represent an important milestone of care for people living with type 1 diabetes (T1D). These devices typically utilise simple control algorithms to select the optimal insulin dose for maintaining blood glucose levels within a healthy range. Online reinforcement learning (RL) has been utilised as a method for further enhancing glucose control in these devices. Previous approaches have been shown to reduce patient risk and improve time spent in the target range when compared to classical control algorithms, but are prone to instability in the learning process, often resulting in the selection of unsafe actions. This work presents an evaluation of offline RL for developing effective dosing policies without the need for potentially dangerous patient interaction during training. This paper examines the utility of BCQ, CQL and TD3-BC in managing the blood glucose of the 30 virtual patients available within the FDA-approved UVA/Padova glucose dynamics simulator. When trained on less than a tenth of the total training samples required by online RL to achieve stable performance, this work shows that offline RL can significantly increase time in the healthy blood glucose range from \(61.6 \pm 0.3\%\) to \(65.3 \pm 0.5\%\) when compared to the strongest state-of-art baseline \((p < 0.001)\). This is achieved without any associated increase in low blood glucose events. Offline RL is also shown to be able to correct for common and challenging control scenarios such as incorrect bolus dosing, irregular meal timings and compression errors. The code for this work is available at: \url{https://github.com/hemerson1/offline-glucose}. 
\end{abstract}

%%Graphical abstract
%\begin{graphicalabstract}
%\includegraphics{grabs}
%\end{graphicalabstract}

%%Research highlights
%\begin{highlights}
%\item Offline reinforcement learning is effective for glucose control in type 1 diabetes.
% \item The approach learns without patient interaction from samples of historical data.
% \item The learned policies outperform the commercial standard of control algorithms.
% \item This method can correct for bolus overestimation and irregular meal schedules.
% \end{highlights}

\begin{keyword}
%% keywords here, in the form: keyword \sep keyword
reinforcement learning \sep type 1 diabetes \sep glucose control \sep artificial pancreas 
%% PACS codes here, in the form: \PACS code \sep code
%% MSC codes here, in the form: \MSC code \sep code
%% or \MSC[2008] code \sep code (2000 is the default)
\end{keyword}

\end{frontmatter}

%% main text ---------------------------------
\section{Introduction}
\label{section:introduction}

% Introduce Type 1 Diabetes and hybrid closed loop systems
Type 1 diabetes (T1D) is an autoimmune disease characterised by an insufficiency of the hormone insulin, which is required for blood glucose regulation. People with T1D must regularly monitor their blood glucose levels and estimate the correct dosage of insulin and carbohydrate intake to avoid dangerous instances of low and high blood glucose. This includes taking bolus insulin to account for ingested meal carbohydrates, in addition to adjusting basal insulin to account for fluctuations between meals. Hybrid closed loop systems provide an opportunity for people with T1D to automatically regulate their basal insulin dosing \cite{DeBock2018EffectProtocol,Abraham2021EffectTrial,Breton2021OneTechnology}. These devices consist of an insulin pump connected to a continuous glucose monitor (CGM) by a control algorithm. The CGM measures the blood glucose level of the user and the control algorithm uses the CGM data to instruct the insulin pump to deliver the required dosage. This process repeats at regular intervals and corrects for deviations in blood glucose from some target blood glucose range or value. Trials of these devices in adults and pediatrics have shown a significant association between the use of hybrid closed loop systems and improvements to time spent in the recommended blood glucose range \cite{McAuley2020SixTrial,Garg2017GlucoseDiabetes,DeRidder2019TheTrials}. Hartnell et al. provides a detailed overview of existing closed loop devices and their functionality \cite{Hartnell2021Closed-loopGuide}.  

%  Briefly discuss the approaches already used PID and the necessity for RL
The majority of commercially available hybrid closed loop systems utilise predictive integral derivative (PID) controllers or model predictive controllers (MPC) \cite{Leelarathna2021Hybrid2021}. These algorithms are robust and easily interpretable, but limit the efficacy of the devices. PID algorithms are prone to overestimating insulin doses following meals and are unable to easily incorporate additional factors which affect blood glucose, such as insulin activity, time of day and exercise \cite{Marchetti2008AnDiabetes,Forlenza2018OngoingStudies}. In contrast, MPCs typically utilise linear or simplified models of glucose dynamics, which are unable to capture the full complexity of the task \cite{Matamoros-Alcivar2021Implementation1,Incremona2018ModelPancreas}. Reinforcement learning (RL) has been proposed as a means of addressing this problem; through which a decision-making agent learns the optimal sequence of actions to take in order to maximise some concept of reward. In glucose control, RL algorithms have demonstrated an ability to learn sophisticated and personalised control policies for individual patients. These policies often outperform their PID and MPC counterparts when trained and evaluated in simulators of glucose dynamics, but are impractical for clinical use in their present state \cite{Myhre2020In-silicoMellitus,Fox2019ReinforcementOpportunities,Fox2020DeepControlb}. Current approaches predominantly utilise online RL algorithms, which require interaction with a patient or simulator during training to develop control policies and learn via a process akin to trial and error. These agents typically start with a poor understanding of their environment and are prone to learning instability \cite{Haarnoja2018SoftActor,Fujimoto2018AddressingMethods}, both of which could feasibly contribute to the selection of dangerous insulin doses. This facet of online RL limits its utility in real-world hybrid closed loop systems; highlighting the necessity for methods capable of learning accurate dosing  policies from clinically obtainable quantities of glucose data without the associated risk.

% Introduce the paper contents
This work presents a proof-of-concept in silico study on the use of offline RL for glucose control, in which an RL agent learns without environmental interaction during training and instead learns from a static dataset of demonstrations collected under another agent. This entails a rigorous analysis of the offline RL algorithms: batch constrained deep Q-learning (BCQ)  \cite{Fujimoto2018Off-PolicyExploration}, conservative Q-learning (CQL) \cite{Kumar2020ConservativeLearning} and twin delayed deep deterministic policy gradient with behavioural cloning (TD3-BC) \cite{Fujimoto2021ALearning} in their ability to develop safe and high performing insulin dosing strategies in hybrid closed loop systems. The presented algorithms are trained and tested across a cohort of 30 virtual patients (10 children, 10 adolescents and 10 adults) and their performance and sample-efficiency is scrutinised with respect to the current strongest online RL and control baselines. Practical limitations such as missing CGM data and suboptimally set PID parameters are also explored, as these are common features of real-world blood glucose data. To ensure the reliable and effective operation of offline RL in the worst-case scenarios, safety is scrutinised across a range of realistic and common control events. This includes overdosing in mealtime insulin, sporadic and irregular meal events and erroneous compression lows caused by force on the CGM insertion site. This work shows that offline RL can yield more effective and safer insulin dosing policies without the patient interaction required by prior RL approaches. Furthermore, this method also utilises significantly smaller samples of data making it more applicable for use in real patients. The presented results provide a foundational overview of the advantages and limitations of using offline RL in glucose control tasks and will provide a reference for future research seeking to integrate offline RL in hybrid closed loop systems before trialling on real-world patients in a controlled setting.  

\section{Related Work}
\label{section:related_work}

% Offline RL in Healthcare
Offline RL is an area of increasing interest in healthcare due to the safety concerns associated with incorrect decision-making \cite{Levine2020OfflineProblems}. Previous healthcare research has focused on developing lung cancer protocols from samples of historical data \cite{Tseng2017DeepCancer:}, identifying optimal recommendations for sepsis treatment \cite{Liu2021OfflineSepsis} and facilitating RL policy evaluation in an offline medical setting \cite{Tang2021ModelSettings}.

% Offline RL in glucose control
Despite the prevalence of offline RL in other domains of medicine, its use in glucose control has been limited \cite{TejedorHernandez2020ControllingFunction}. Javad et al. used a q-learning algorithm trained offline on samples of clinical data to select the optimal daily basal dose for patients of a given demographic \cite{Javad2019AStudy}. Shi et al. presented an offline q-learning approach trained on the OhioT1DM dataset for selecting discrete doses of basal insulin at hourly intervals \cite{Shi2022StatisticallyHorizons}. Similarly, Li et al. used a model-based RL algorithm to learn the blood glucose dynamics of patients recovering from diabetic ketoacidosis \cite{Li2022ElectronicOptimizing}. This learned model was then used to train an online q-learning algorithm to select the optimal basal dose at three hour intervals over a 24-hour period. Although these approaches presented methods for learning dosing policies offline, the timeframe over which they act would be insufficient for managing the short-term blood glucose fluctuations associated with meals and exercise. The most significant application of offline RL for hybrid closed loop systems was presented in Fox, in which the methods BCQ and bootstrapping error accumulation reduction (BEAR) were compared to online RL approaches trained on an imperfect simulator of a single adult patient \cite{Fox2020MachineManagement}. Analysis showed that although offline RL was capable of developing competent control policies, the achieved performance was less than that obtained using an online approach trained on an imperfect simulator. 

% Blood glucose modelling and supervised learning
Blood glucose forecasting is an integral aspect of T1D management \cite{Woldaregay2019Data-drivenDiabetesb}, consequently significant research has been focused on extending established control algorithms using blood glucose prediction models dervied via supervised learning. This has included using quantile regression to predict upper and lower bounds on future blood glucose values \cite{Dutta2018RobustNetworksb}, bolstering fuzzy logic controllers with neural network prediction \cite{Allam2013BloodController} and leveraging a pair neural networks to forecast blood glucose and select insulin doses based on the predictions \cite{FernandezdeCanete2012ArtificialDiabetes}. A number of these approaches have also been experimentally validated within animal trials \cite{Bahremand2019NeuralSystemb, Kirilmaz2022ARats}. Of particular note, Chen et al. trains a behavioural cloning agent on the insulin doses of an MPC demonstrator; providing a sample-efficient alternative of learning insulin dosing strategies without the risk of dangerous action selection during training. Supervised learning has also been used in hybrid closed loop systems for functions secondary to insulin dosing, such as hypoglycemia prediction \cite{Mujahid2021MachineChallenges} and detecting unreported carbohydrate consumption \cite{Mosquera-Lopez2023EnablingIntelligence}. 

% Methods for safely learning glucose control policies with patient interaction
Within online RL, several attempts have been made to address concerns around safety and learning instability in glucose control. Fox et al. employed a transfer learning approach to develop dosing policies from a general patient population before fine-tuning them on target patients \cite{Fox2020DeepControlb}. Lim et al. used a PID controller to guide an online RL algorithm in the early stages of its learning; progressively introducing a greater proportion of RL agent actions \cite{Lim2021AValidation}. Zhu et al. developed an online RL algorithm capable of integration with a dual hormone pump, allowing control of both insulin and glucagon dosing and hence for low blood glucose corrections to be made through glucagon infusion \cite{Zhu2021BasalValidation}. These approaches all showed comparable or reduced time spent in the potentially dangerous low blood glucose range when contrasted with MPC and PID algorithms, but are limited by their reliance on glucose dynamics simulators. These simulators represent a simplification of true blood glucose dynamics; with almost all unable to incorporate common events such as exercise, stress and illness. In moving towards use in real-world hybrid closed loop systems, online RL performance is likely to deteriorate as latent variables start to influence the environment \cite{Zhao2020Sim-to-RealSurvey}. In addition, glucose dynamics simulators allow for unlimited training data generation. Of the approaches highlighted which explicitly stated their sample size, the algorithms used 7,000 days ($>$19 years) \cite{Fox2020DeepControlb} and 1,530 days ($>$4 years) \cite{Zhu2021BasalValidation} of simulated data to develop personalised dosing policies. In an in vivo setting, allowing an RL algorithm to control a patient's blood glucose for several years without any associated guarantee of safety would be unethical. 

% Explain the distinctions between this work and others
This work represents the first rigorous evaluation of offline RL for glucose control; contrasting the performance of a diverse range of state-of-the-art algorithms in a comprehensive cohort of virtual patients. Evaluation is performed for a combined 41,660 virtual days ($>$114 years) and the presented algorithms are evaluated inline with the current clinical guidance for assessing patient glucose control. This analysis holds a particular focus on exploring the practical and safety limitations of offline RL within hybrid closed loop systems. In silico evaluation of this nature is essential in justifying trials of offline RL for glucose control in real patient populations.

\section{Materials and Methods}
\label{section:material_methods}

\subsection{Problem Formulation}

% Introduce POMDP & discuss the state, action and reward required for this environment
The task of glucose control in hybrid closed loop systems can be modelled as a partially observable Markov decision process (POMDP) described by (\(\mathcal{S}\), \(\mathcal{A}\), \(\Omega\),  \(\mathcal{T}\), \(\mathcal{R}\), \(\mathcal{O}\)). In each timestep \(t\), the agent in a single state \(s \in \mathcal{S}\), describing the current environment, interacts with the environment via an action \(a \in \mathcal{A}\) and receives a reward \(r = \mathcal{R}(s) \in \mathbb{R}\) specifying the optimality of the action, before transitioning into a new state \(s' \in \mathcal{S}\) with transition probability \(P(s'|s, a) = \mathcal{T}(s, a, s')\). In a POMDP, the agent is unable to directly view the next state \(s' \in \mathcal{S}\), and instead receives an observation \(o \in \Omega\) determined by probability \(P(o'|s', a) = \mathcal{O}(o', s', a)\) which provides insight into the next state \cite{Omidshafiei2017DeepObservability}. In the glucose control setting, the state and action are defined by the patient blood glucose value \(g_t\) and by the basal insulin dose \(i_t\) in that timestep. The partial observability of the state results from the inherent noise in the CGM devices used to make blood glucose measurements and the dependency of blood glucose values on historical data, such as ingested carbohydrates \(c_t\), bolus doses \(b_t\) and previous blood glucose values \cite{Xie2018ReductionPolymer}. Contextual information was incorporated in this work by utilising the following state:
\begin{equation}
    s_t = [g_{t}, g_{t-1}, g_{t-2}, ..., g_{t-8}, I_t, M_t].
\end{equation} 
This representation consists of a rolling window of measured blood glucose values updated every three minutes using the past four hours of blood glucose data spaced at 30-minute intervals, with $g_t$ being the blood glucose in the current timestep and and $g_{t-8}$ being the blood glucose four hours prior. An estimation of the combined insulin activity of basal and bolus insulin (insulin-on-board) \(I_t\) and an estimation of carbohydrate activity \(M_t\) are also included and given by \cite{Toffanin2013DynamicVariation}:
\begin{equation}
    I_t = \sum^{N}_{t'=0} \left( 1 - \frac{t'}{N+1} \right)\left[b_{t - t'} + i_{t - t'}\right], \hspace{0mm}
\end{equation} \break
\begin{equation}
    M_t = \sum^{N}_{t'=0} \left( 1 - \frac{t'}{N+1} \right) \left[ c_{t - t'} \right],
\end{equation}
\noindent
where $N$ represents the number of prior timesteps the algorithm considers in its decision-making. This representation simplifies true insulin and carbohydrate activity by assuming that they decay linearly to zero over a four hour period. This state was selected in place of the full sequence of blood glucose, carbohydrate and insulin data utilised in other approaches to reduce state dimensionality and to avoid modifying the offline RL methods to incorporate recurrency \cite{Fox2020DeepControlb,Fox2019ReinforcementOpportunities,Zhu2021BasalValidation}. The reward for the agent was given by the negative of the Magni risk function, which models the clinical risk for a given blood glucose value \cite{Kovatchev1997SymmetrizationApplications}. An additional penalty of -1e\textsuperscript{5} was added for blood glucose values beyond the physiologically feasible range of 10 to 1,000 mg/dl. This modification is several orders of magnitude larger than the obtainable reward under the Magni risk function and was included as an incentive for the agent to not purposefully terminate the environment and avoid future negative reward. The parameters for the risk function are given as follows \cite{Fox2020DeepControlb}:
\begin{equation}
    risk(g_t) = 10 \cdot \left(3.5506 \cdot \left( \log \left(g_t \right)^{0.8353} - 3.7932 \right) \right)^2.
\end{equation}
A reward function of this form ensures that low blood glucose events are punished more severely than high blood glucose events; reflecting the greater immediate risk low blood glucose events pose to patient health. The reward is at a maximum when blood glucose is approximately in the centre of the target range (70-180 mg/dl). This reward function was selected as prior work found it resulted in the greatest empirical performance \cite{Fox2020DeepControlb}, however alternative reward functions for glucose control are referenced in Tejedor et al. \cite{TejedorReinfocementLearningReview2020}.

\subsection{Offline Reinforcement Learning}

% Introduce the offline RL algorithms in this paper
RL algorithms learn the optimal series of actions to take in a given environmental state to maximise the agent's total reward received. This state-action mapping is referred to as the agent's policy and is updated from demonstrations of interactions with the chosen environment. Typically, environmental demonstrations are generated in an online manner, in which the agent takes actions in the environment and updates its understanding in parallel. However in offline RL, the agent is incapable of environmental interaction during the training procedure and instead must rely on samples generated by a demonstrator in a retrospective or simulated dataset, such as a PID algorithm, to build an understanding of the POMDP \cite{Levine2020OfflineProblems}. RL methods can be broadly divided into model-free and model-based, which are distinguished by the use of a dynamics model through which the transition probability is approximated. Alternatively, model-free approaches often estimate the Q-function, which defines the expected future reward of the agent when taking action \(a\) in state \(s\) and then continuing to make decisions using the learned policy \(\pi(s)\):  
\begin{equation}
Q^\pi = \left[\sum^\infty_{i=t+1} \gamma^i r(s_i, a_i, s_{i+1}) \mid s, a \right],
\end{equation}
where \(\gamma \in [0, 1)\) weights the agent's future reward. In the simplest form, an agent updates its approximation of the Q-function via the Bellman equation:
\begin{equation}
Q(s_t, a_t) \leftarrow (1 - \alpha) \cdot Q(s_t, a_t) + \alpha \cdot \left(r_t + \gamma \cdot \max_{a} Q(s_{t+1}, a) \right),
\label{eq:bellman_update}
\end{equation}
where \(\alpha \in (0, 1]\) is the learning rate. The agent then utilises the learned Q-function to update its policy, for example the online RL algorithm DQN selects the action in each state corresponding to the maximum expected future reward \cite{Mnih2013PlayingLearning}:
\begin{equation}
\pi(s) = \argmax_a Q(s, a).
\end{equation}
A central challenge of applying offline RL to real-world tasks is distributional shift, in which an offline RL agent encounters states significantly different from those observed in the training data \cite{Kumar2020ConservativeLearning}. In these instances, extrapolation to out-of-distribution states can result in the erroneous overestimation of the Q-function and the selection of poor actions. This is particularly significant in safety-focused tasks such as glucose control. This work applies the following model-free offline RL approaches to reduce Q-function overestimation:

\begin{itemize}
\setlength\itemsep{2mm}

\item \textbf{Batch Constrained Deep Q-learning (BCQ)} modifies the DQN algorithm by constraining the agent to select similar actions and states to those observed in the training data \cite{Fujimoto2018Off-PolicyExploration}. In addition to the Q-function estimator, a variational auto-encoder is trained to generate similar actions to those in the training data when in a given state and diversity is incorporated by adjusting the generated actions using a perturbation model. State visitation is also constrained by applying a modified version of clipped double q-learning, whereby two separate Q-function approximators are updated using the minimum of their estimates \cite{Fujimoto2018AddressingMethods}. This adjustment reduces overestimation bias by minimising the Q-function in each update. As a consequence of this reduction, high Q-values are preferentially assigned to states with low variance that have high visitation in the training dataset.

\item \textbf{Conservative Q-learning (CQL)} expands on prior works, such as BCQ, by employing an alternative approach of addressing Q-function overestimation
on out-of-distribution state-action tuples. CQL learns a conservative Q-function, which acts as a lower bound on the true Q-value for a given policy \cite{Kumar2020ConservativeLearning}. This is incorporated by modifying the traditional Q-function update to include an additional term which simultaneously minimises Q-value estimates on unseen state-action tuples, while maximising the Q-values of tuples observed in the dataset. In theory, this change encourages the agent to preferentially select in-distribution actions by assigning them high Q-value estimates.    

\item \textbf{Twin Delayed DDPG with Behavioural Cloning (TD3-BC)} modifies the established off-policy online RL method TD3, to include a behavioural cloning term in its loss; encouraging the policy to select actions observed in the training distribution \cite{Fujimoto2021ALearning}. Beyond the aforementioned modification, TD3 utilises a number of methods to avoid Q-function overestimation such as clipped double Q-learning and Q-function smoothing.  

\end{itemize}

\subsection{Baselines}

% Introduce PID and SAC-RNN
The performance of the offline RL methods were compared to two baseline algorithms: a tuned PID controller and the online RL method, recurrent soft actor-critic (SAC-RNN). The PID algorithm is a robust classical control mechanism capable of correcting for both short and long-term deviations from a target blood glucose value \(g_{\text{target}}\) and operates in each timestep $t$ according to \cite{Ngo2018ControlAlgorithm}: 
\begin{equation}
    i_t = k_p \cdot \left( g_{\text{target}} - g_t \right) + k_i \cdot \sum^{t}_{t'=0}\left[ g_{t'} - g_{target} \right] + k_d \cdot \left( g_t - g_{t-1} \right),
\end{equation}
where \(k_p\), \(k_i\) and \(k_d\) are parameters to be set. To ensure the strongest comparison, the parameters were personalised to each patient and were selected using a grid-search method to maximise the reward collected over a 10-day test period. This algorithm is one of the most well used in hybrid closed loop systems \cite{TejedorReinfocementLearningReview2020}, including the Medtronic 670g and 780g Guardian 3 sensors \cite{Leelarathna2021Hybrid2021b}.   

The SAC-RNN method represents one of the state-of-the-art algorithms in glucose control and was recently presented in Fox et al. \cite{Fox2020DeepControlb}. The method used in this paper is a variation of this implementation, using long-short-term memory (LSTM) layers in place of gated recurrent unit (GRU) layers. This substitution was made as it was empirically found to improve algorithmic performance and policy stability and thus provided a stronger baseline. Futher details of the implementation can be found in the provided repository. 

\subsection{Glucose Dynamics Simulation}

% Discuss the UVA/Padova Simulator and the modifications for this experiment
The UVA/Padova T1D glucose dynamics model was used to generate data in this work, as it allowed control over the size and quality of the training dataset \cite{Xie2018SimglucoseV0.2.1}. In addition, to providing a rigorous platform for evaluating the developed control algorithms without relying on unverified offline evaluation methods. This software simulates the human metabolic system using dietary models of glucose-insulin kinetics and is designed as a substitute for pre-clinical trials in the development and testing of T1D control algorithms \cite{DallaMan2014TheFeatures}. The simulator can model a cohort of 30 virtual patients (10 children, 10 adolescents and 10 adults) and their individualised responses to meal carbohydrates, basal/bolus insulin dosing and interaction with CGM/pump devices. For the purpose of this study, all virtual patients utilised a CGM with a three-minute sampling rate joined to a pump device. This sampling rate is the default for the simulator and is within the range of real-world systems (1 to 15 minutes) \cite{Bergenstal2018UnderstandingData}. Few physiological processes and events relevant to T1D management occur over timeframes shorter than the CGM sampling rate, therefore control algorithms should perform comparably across the realistic range. Continuously valued basal insulin doses were selected by the RL/PID agent with bolusing administered using the following controller \cite{Schmidt2014BolusCalculators}:
\begin{equation}
    I_t = \frac{c_t}{CR} + \left( \sum^{59}_{t'=0}c_{t -t'} = 0 \right) \cdot \left[ \frac{g_t - g_{\text{target}}}{CF} \right],
\end{equation}
where \(g_{\text{target}} =\) 144 mg/dl is the target blood glucose level and corresponds to the greatest reward in the Magni risk function, where the probability of high and low blood glucose events are at a minimum \cite{Magni2007ModelTrial}. In addition, \(CF\) and \(CR\) are patient-specific parameters and were chosen from Fox et al. \cite{Fox2020DeepControlb}. Three meals and three snack events were included in the simulator with the time and quantity of carbohydrate ingestion for each event being modelled by a normal distribution.

\subsection{Experimental Setup}

% Data Collection and Training
\subsubsection{Data Collection and Training} Each algorithm was trained on \(1\text{e}^5\) samples of data collected over epochs of 10 days; the equivalent of 208 days of glucose data. Samples were collected from 30 simulated patients (10 children, 10 adolescents and 10 adults) and individually each patient's data was used to train each algorithm across three seeds. Training and testing in adolescent and child populations is important for the adoption of T1D technology as these groups are typically more susceptible to high blood glucose events and increased glucose variability \cite{Miller2015CurrentRegistry}. The training data was collected using a PID algorithm tuned to achieve the maximum reward over a 10-day test period and noise was added to improve exploration within the generated dataset. Basal noise was introduced by using an Ornstein-Uhlenbeck process \cite{Bibbona2008TheNoise}. In addition, bolus noise was introduced by adding a 10\% estimation error to the simulated carbohydrate intake; this more closely models the uncertainty observed in patient calculations of bolus doses. The hyperparameters used for the offline RL algorithms were unchanged from their original implementations. This choice was made as hyperparameter optimisation would require patient interaction via the simulator to validate model performance; potentially harming the participant in the process. Hyperparameter selection could also be performed using offline evaluation methods however this is outside the scope of this work.  

\begin{table}[H]
\footnotesize
\centering
% Vertical and Horizontal Padding
\setlength{\tabcolsep}{0.5em}
\def\arraystretch{1.5}

\begin{tabular}{p{2.0cm} p{8.0cm} >{\centering\arraybackslash}p{2.0cm} }
\toprule
Metric & Description & Target \\
\midrule
Time-in-Range (TIR) & The percentage time for which blood glucose measurements fall within the healthy glucose range (70-180 mg/dl). Increased TIR is strongly associated with a reduced risk of developing micro-vascular complications  \cite{Beck2019ValidationTrials}. & $>$70\%\mbox{*} \cite{Battelino2019ClinicalRange} \\

Time-Below-Range (TBR) & The percentage time for which blood glucose measurements fall in the low blood glucose range ($<$70 mg/dl). Combined with TIR, this can additionally act as an indirect measure of time spent in the high blood glucose range ($>$180 mg/dl). & $<$4\% \cite{Battelino2019ClinicalRange} \\

Coefficient of Variation (CV) & The relative dispersion of blood glucose values around their mean. Increased CV is linked to an elevated risk of severe low blood glucose events ($<$54 mg/dl) and vascular tissue damage \cite{Ceriello2019GlycaemicImplications}. & $<$36\% \cite{Danne2017InternationalMonitoring}\\

Failure & The percentage of test rollouts in which blood glucose levels reached values $<$10 mg/dl or $>$1000 mg/dl. For context, blood glucose $<$40 mg/dl is considered life-threatening and can result in major cardiovascular and cerebrovascular problems \cite{Kalra2013Hypoglycemia:Complication}. & 0\% \\
\bottomrule
\end{tabular}
\caption{A description of the glycemic control metrics utilised for evaluation alongside their clinically recommended values. \mbox{*}For ages $<$25 this target decreases to $>$60\%.}
\label{tab:glucoseMetrics}
\end{table}

% Evaluation
\subsubsection{Evaluation} Performance was evaluated by monitoring blood glucose levels over a simulated 10-day test period and aggregating the results over three test seeds per training seed to ensure sufficient variation in test scenarios. The metrics utilised for evaluation are given in Table \ref{tab:glucoseMetrics} and were used in addition to the sum of reward for assessing algorithmic performance. Friedman rank tests and Wilcoxon signed-rank tests were used to assess significance between control algorithm outcomes. In addition, the standard error between test seeds is presented alongside each measurement. Further tests were employed to identify the practical limitations of using offline RL algorithms in hybrid closed loop systems. This included evaluation on datasets with: 1) decreasing sample size, 2) demonstrations from suboptimal PID demonstrators and 3) sequences of missing CGM data caused by temporary sensor transmitter errors for extended periods. In addition, the potential of offline RL for safer blood glucose management was explored by engineering several common and challenging control scenarios. These included patients with: 4) consistently overestimated bolus insulin for meals, 5) irregular meal schedules with greater uncertainty in meal times and 6) frequent erroneous low blood glucose readings caused by compression lows. Compression lows result from pressure on the CGM sensor insertion site and are caused by the redistribution of the interstitial fluid from which blood glucose is measured \cite{Mensh2013SusceptibilityPosition}.  

\section{Results}
\label{section:results}

\subsection{Offline Reinforcement Learning vs. Baseline Control Methods}

A comparison of the described offline RL methods with baseline approaches is detailed in Table \ref{tab:mainResults}. Of the methods presented, the offline RL algorithm TD3-BC achieved the best performance; obtaining the greatest reward and hence the lowest Magni risk over the evaluation period. In addition, the algorithm yields a \(3.7 \pm 0.6\%\) increase to TIR and a reduction in TBR when compared to the PID algorithm. The observed difference may in part be due to the inclusion of carbohydrate information in the state; providing an early indication of when a sharp rise in blood glucose may occur. The ability of RL algorithms to readily incorporate new sensor modalities without explicit programming is a significant advantage of the approach over non-machine learning based methods and could be utilised to incorporate a patient's individualised response to exercise or stress if provided with the relevant sensor data. 

This equates to almost an additional hour per day in which patients would experience an improved quality of life. BCQ also shows a similar level of improvement to TIR, however this is coupled with a \(0.6 \pm 0.1\%\) increase to TBR. The use of BCQ also resulted in an increase to CV of \(1.6 \pm 0.4\%\). This value has been found clinically to fall within the region of 31.0\% to 42.3\% for people with T1D \cite{Monnier2017TowardDiabetes}. An elevated CV should indicate an increased risk of low blood glucose events, which is evidenced with the BCQ algorithm. The origin of the difference is most likely the use of the Magni risk function for defining reward, as this value does not consider blood glucose variability within its risk calculation. 

% Combined Results for Offline vs Online ----------------------------------------
\begin{table}[H]
\footnotesize
\centering
\setlength{\tabcolsep}{0.5em}
\begin{tabular}{cccccc}
\toprule
Algorithm & Reward & TIR (\%) & TBR (\%) & CV (\%) & Failure (\%) \\
\midrule
BCQ\textsuperscript{$\dagger$} & -41,034 $\pm$ 1,060 & \textbf{65.8 $\pm$ 0.6} & 1.0 $\pm$ 0.1 & 35.1 $\pm$ 0.4 & \textbf{0.00} \\
CQL\textsuperscript{$\dagger$} & -45,259 $\pm$ 1,071 & 56.2 $\pm$ 0.5 & \textbf{0.1 $\pm$ 0.1} & 30.3 $\pm$ 0.3 & \textbf{0.00} \\
\textbf{TD3-BC}\textsuperscript{$\dagger$} & \textbf{-37,955 $\pm$ 547} & 65.3 $\pm$ 0.5 & 0.2 $\pm$ 0.1 & 33.3 $\pm$ 0.2 & \textbf{0.00} \\
SAC-RNN\textsuperscript{$\ddagger$} & -93,480 $\pm$ 71,826 & 34.9 $\pm$ 3.1 & 4.1 $\pm$ 0.7 & \textbf{29.6 $\pm$ 1.3} & 13.3 \\
PID\textsuperscript{$\mathsection$}  & -49,077 $\pm$ 556 & 61.6 $\pm$ 0.3 & 0.4 $\pm$ 0.1 & 33.5 $\pm$ 0.2 & \textbf{0.00} \\
\bottomrule
\end{tabular}
\caption{The mean performance of the offline RL algorithms: BCQ, CQL and TD3-BC against the online RL approach SAC-RNN and the control baseline PID. TD3-BC can be seen to significantly improve the proportion of TIR when compared to the PID and the SAC-RNN algorithms. This is done so without any associated increase in risk (reward) or TBR. Statistical significance was confirmed via a Friedman rank test for all glucose metrics \((p < 0.05)\). $\dagger$, $\ddagger$ and $\mathsection$ indicate an offline RL, online RL and classical control algorithm respectively, with the best performing algorithm highlighted in bold.}
\label{tab:mainResults}
\end{table}
% ----------------------------------------------------------------------------

The online RL algorithm SAC-RNN, performs comparably worse than the PID and offline RL approaches in almost all metrics; terminating the environment and thus harming the patient in 13.3\% of the test rollouts. In this instance, CV does fall well within the recommended threshold of $<$36\%, however this is likely a consequence of patients having high blood glucose for \(61.0 \pm 3.2\%\) of the evaluation period and therefore being closer to their blood glucose equilibrium point \cite{Danne2017InternationalMonitoring}. The performance of SAC-RNN in this work significantly differs from the results obtained previously under similar implementations \cite{Fox2020DeepControlb,Viroonluecha2021EvaluationLearning}. This difference is most likely a result of differing evaluation methods. In this work, SAC-RNN was trained for the full duration and evaluated using the resulting weights, however in previous implementations performance was measured on a validation environment after each episode and the highest performing weights were selected for evaluation. One such online method reported that using the final weights in place of the best performing weights resulted in almost half of test rollouts ending in termination \cite{Fox2020DeepControlb}. This work elected to use the final weights for online evaluation, as it was concluded to be more indicative of how the algorithm would be utilised in a patient setting. Whereby, the algorithm would seek to continually adapt to changes in the patient's lifestyle and blood glucose dynamics.   

% Results per Group -------------------------------------------------------------

\subsection{Offline Reinforcement Learning Performance by Patient Age}

% Adults ----------------

\begin{table}[H]
\footnotesize
\setlength{\tabcolsep}{0.5em}
\begin{subtable}[t]{1.0\textwidth}
\centering
\caption{Adults (aged 26 to 68 years).}
\begin{tabular}{cccccc}
\toprule
Algorithm & Reward & TIR (\%) & TBR (\%) & CV (\%) & Failure (\%) \\
\hline
\textbf{BCQ}\textsuperscript{$\dagger$} & \textbf{-17,445 $\pm$ 290} & \textbf{72.6 $\pm$ 0.6} & 0.1 $\pm$ 0.0 & 25.6 $\pm$ 0.2 & \textbf{0.00} \\
CQL\textsuperscript{$\dagger$} & -20,748 $\pm$ 301 & 62.5 $\pm$ 0.5 & \textbf{0.0 $\pm$ 0.0} & \textbf{22.9 $\pm$ 0.2} & \textbf{0.00} \\
TD3-BC\textsuperscript{$\dagger$} & -19,538 $\pm$ 381 & 70.0 $\pm$ 0.6 & 0.1 $\pm$ 0.1 & 26.0 $\pm$ 0.2 & \textbf{0.00} \\
SAC-RNN\textsuperscript{$\ddagger$} & -49,600 $\pm$ 6,075 & 42.5 $\pm$ 3.4 & 5.1 $\pm$ 0.1 & 26.0 $\pm$ 1.5 & 6.6 \\
PID\textsuperscript{$\mathsection$}  & -19,783 $\pm$ 262 & 65.8 $\pm$ 0.3 & \textbf{0.0 $\pm$ 0.0} & 24.4 $\pm$ 0.2 & \textbf{0.00} \\
\bottomrule
\end{tabular}
\label{tab:mainResultsPerGroupAdult}
\end{subtable}

% Adolescents ----------------

\begin{subtable}[t]{1.0\textwidth}
\centering
\caption{Adolescents (aged 14 to 19 years).}
\begin{tabular}{cccccc}
\toprule
Algorithm & Reward & TIR (\%) & TBR (\%) & CV (\%) & Failure (\%) \\
\hline
BCQ\textsuperscript{$\dagger$} & -39,932 $\pm$ 347 & \textbf{64.9 $\pm$ 0.3} & 1.6 $\pm$ 0.1 & 33.28 $\pm$ 0.3 & \textbf{0.00} \\
CQL\textsuperscript{$\dagger$} & -43,847 $\pm$ 389 & 56.12 $\pm$ 0.4 & \textbf{0.1 $\pm$ 0.0} & 27.6 $\pm$ 0.3 & \textbf{0.00} \\
\textbf{TD3-BC} \textsuperscript{$\dagger$} & \textbf{-39,363 $\pm$ 614} & 62.0 $\pm$ 0.6 & \textbf{0.1 $\pm$ 0.1} & \textbf{26.1 $\pm$ 0.3} & \textbf{0.00} \\
SAC-RNN\textsuperscript{$\ddagger$} & -80,102 $\pm$ 5,597 & 25.9 $\pm$ 3.2 & 2.7 $\pm$ 0.9 & \textbf{26.1 $\pm$ 0.2} & 13.3 \\
PID\textsuperscript{$\mathsection$}  & -40,180 $\pm$ 465 & 60.6 $\pm$ 0.2 & 0.1 $\pm$ 0.0 & 30.75 $\pm$ 0.2 & \textbf{0.00} \\
\bottomrule
\end{tabular}
\label{tab:mainResultsPerGroupAdolescents}
\end{subtable}

% Children ----------------

\begin{subtable}[t]{1.0\textwidth}
\centering
\caption{Children (aged 7 to 12 years).}
\begin{tabular}{cccccc}
\toprule
Algorithm & Reward & TIR (\%) & TBR (\%) & CV (\%) & Failure (\%) \\
\hline
BCQ\textsuperscript{$\dagger$} & -61,374 $\pm$ 2,543 & 56.9 $\pm$ 0.9 & 1.2 $\pm$ 0.2 & 41.9 $\pm$ 0.6 & \textbf{0.00} \\
CQL\textsuperscript{$\dagger$} & -66,346 $\pm$ 2,522 & 44.2 $\pm$ 0.7 & \textbf{0.3 $\pm$ 0.1} & 37.2 $\pm$ 0.4 & \textbf{0.00} \\
\textbf{TD3-BC} \textsuperscript{$\dagger$} & \textbf{-51,713 $\pm$ 646} & \textbf{60.1 $\pm$ 0.4} & \textbf{0.3 $\pm$ 0.1} & 40.4 $\pm$ 0.2 & \textbf{0.00} \\
SAC-RNN\textsuperscript{$\ddagger$} & -97,760 $\pm$ 6,339 & 37.3 $\pm$ 2.8 & 4.0 $\pm$ 1.0 & \textbf{36.0 $\pm$ 1.4} & 20.0 \\
PID  & -57,700 $\pm$ 941 & 54.2 $\pm$ 0.4 & 1.7 $\pm$ 0.1 & 41.6 $\pm$ 0.3 & \textbf{0.00} \\
\bottomrule
\end{tabular}
\label{tab:mainResultsPerGroupChild}
\end{subtable}

\caption{The mean glucose control performance for the individual patient models divided by age group, as specified in the UVA/Padova T1D Simulator. TD3-BC represents the only RL approach to perform better than the PID across the three patient groups. The greatest improvement is observed in adults where BCQ and TD3-BC yield a significant increase to time in the target range, equivalent to an improvement of 100 mins/day and 60 mins/day. Statistical significance was confirmed for all metrics via a Friedman rank test \((p < 0.05)\). $\dagger$, $\ddagger$ and $\mathsection$ indicate an offline RL, online RL and classical control algorithm respectively, with the best performing algorithm highlighted in bold.}
\label{tab:mainResultsPerGroup}
\end{table}

Table \ref{tab:mainResultsPerGroup} presents a breakdown of glucose control performance when divided by patient age. As in Table \ref{tab:mainResults}, the offline RL algorithm TD3-BC performs the most consistently across the patient groups, achieving a greater reward to the PID across all categories. The greatest improvements to TIR are observed in adult patients, where BCQ and TD3-BC achieve an increase of \(6.8 \pm 0.7\%\) and \(4.2 \pm 0.7\%\) respectively. This observed difference is significant enough to push TIR to within the recommended margin for that age group, which if sustained would potentially lead to markedly better long-term health outcomes for that population. The results in the child group are also particularly promising, whereby the TD3-BC approach yields a \(5.9 \pm 0.4\%\) increase to TIR and a \(1.4 \pm 0.1\%\) reduction to TBR. Children represent one of the most challenging control groups within the T1D simulator and in real life, as evidenced by the significantly lower reward and greater glycemic variability obtained under the PID in that group. This control disparity is predominately due to differing insulin sensitivities between the age groups. Insulin sensitivity has been identified to negatively correlate with a patient's age and consequently smaller doses of insulin elicit greater blood glucose responses in children and adolescents and require greater precision in insulin dosing \cite{Szadkowska2008InsulinAdolescents}. Achieving an improvement of this magnitude is encouraging for the transition of offline RL algorithms to real patient data in which blood glucose relationships are likely to be more complex and depend on a greater number of environmental factors. The TD3-BC algorithm was selected for further evaluation due to the high performance the approach achieved across the 30 virtual patients, in addition to its consistent safety profile.  

% -----------------------------------------------------------------------------------

% Further Experiments -----------------------------------------

\subsection{Implementation Challenges of Offline Reinforcement Learning in Glucose Control}
\label{sec:implementation_experiments}

\begin{table}[H]
\footnotesize
\centering
% Vertical and Horizontal Padding
\setlength{\tabcolsep}{0.5em}
\def\arraystretch{1.5}

\begin{tabular}{p{2.5cm} p{4.5cm} p{5.0cm}}
\toprule
Experiment & Motivation & Description \\
\midrule
1) Sample Size & Patients are unlikely to adhere to lengthy periods of data collection. & Datasets of size: 1e\textsuperscript{4}, 5e\textsuperscript{4},  1e\textsuperscript{5} and 5e\textsuperscript{5} were used to train TD3-BC for fixed episodes.\\

2) Suboptimal Demonstrations & Patient insulin requirements evolve due to physiological and lifestyle changes, therefore PID parameters are unlikely to always be optimal. & The PID parameters corresponding to the 10\textsuperscript{th} and 20\textsuperscript{th} greatest reward were used as the demonstrator for data collection.\\

3) Missing Data & Interruptions in CGM sensor-transmitter communication commonly lead to intermittent drops in blood glucose readings \cite{Drecogna2021DataData}. & The CGM would once a day (1/500) and twice a day (1/250) fail to record blood glucose measurements for at most 30 minutes.\\

4) Meal Overestimation & Carbohydrate estimation is a challenging task and errors often occur \cite{Meade2016AccuracyAdults}. & All carbohydrate consumption was overestimated by a mean of 20\% and 40\%.\\

5) Irregular Meal Schedules & Irregular meal schedules are correlated with worse glycemic control \cite{Ahola2019MealControl}. & Meal time standard deviation was increased from 0 to 30 to 60 minutes.\\ 

6) Compression Error & Erroneous drops in glucose readings of as much as 25 mg/dl can occur when pressure is applied to a CGM device \cite{Mensh2013SusceptibilityPosition}. & The CGM would once a day (1/500) or twice a day (1/250) record blood glucose a maximum of 30 mg/dl lower for a duration of at most 30 minutes.\\

\bottomrule
\end{tabular}
\caption{Overview of the additional implementation and safety experiments performed to address challenges around the practical use of offline RL algorithms in real-world hybrid closed loop systems.}
\label{tab:further_exp_details}
\end{table}

Experiments in this section explore glucose control specific challenges which may undermine the utility of offline RL in the real-world. The full implementation details for the additional experiments are described in Table \ref{tab:further_exp_details}. The selected TD3-BC algorithm was trained on a single NVIDIA GeForce RTX 2080 Ti GPU and an Intel Core i9-9900K CPU at 3.60 GHz for a duration of approximately 10 minutes per 1e\textsuperscript{5} samples of glucose data. 

\begin{figure}[H]
\hspace{-7mm}
\centering
\includegraphics[width=0.85\textwidth]{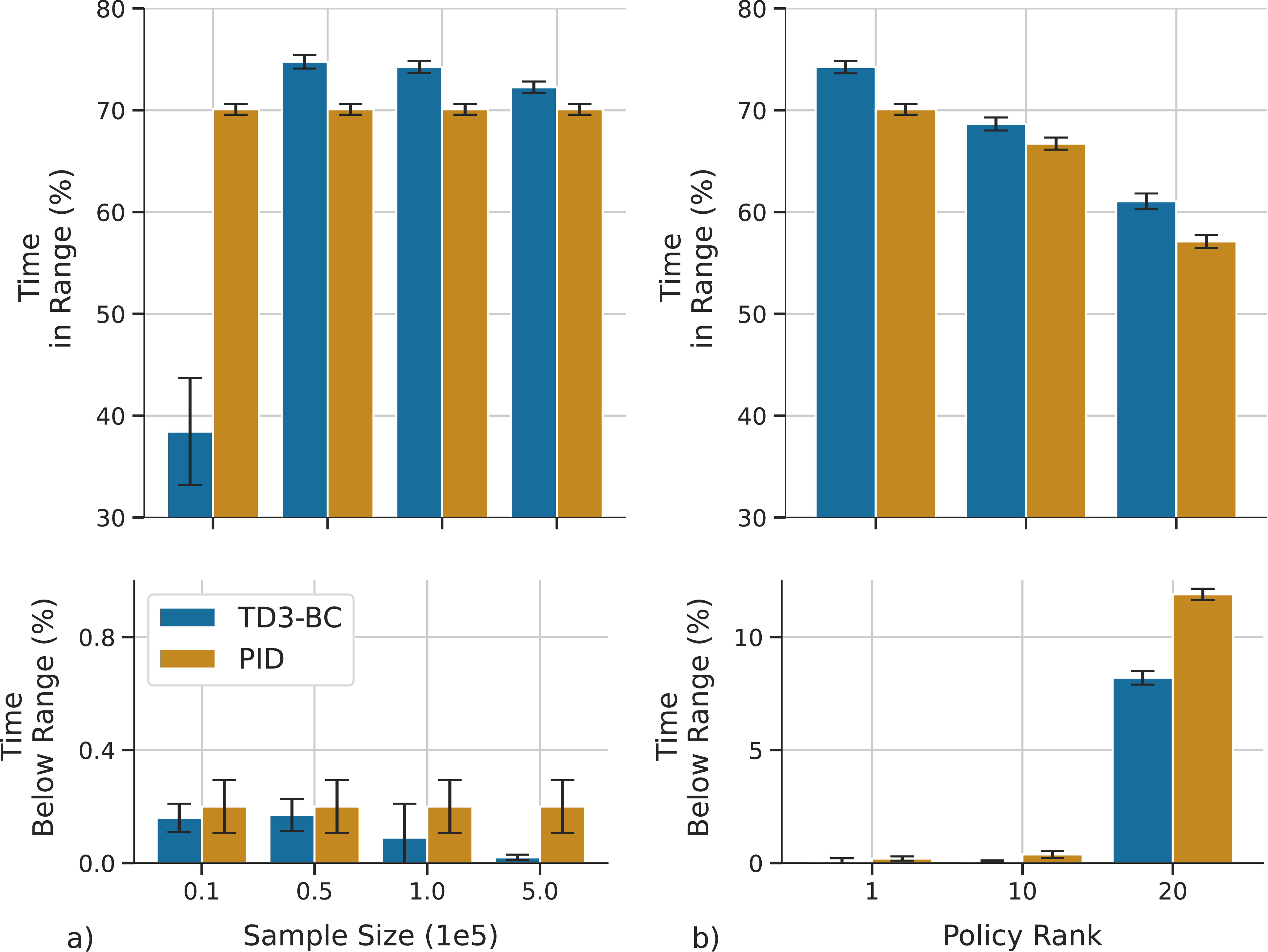}
\caption{a) TD3-BC and PID performance for a varying number of training samples. The offline RL approach TD3-BC can be seen to surpass PID performance when trained on at least 5e\textsuperscript{4} samples of glucose data (100 days of glucose data). With greater sample size, TD3-BC yields an increase to time in the target glucose range (TIR) with an accompanying decrease to the proportion of low blood glucose events (TBR).  b) TD3-BC and PID performance when trained on expert demonstrators of different ability. TD3-BC can be seen to significantly improve TIR, while achieving comparable or better TBR for all training policies. The reduction in TBR is largest when trained on the 20\textsuperscript{th} best policy. Wilcoxon signed-rank tests confirmed the significance of the TD3-BC TIR improvement across all sample sizes and demonstrator abilities (\(p < 0.05\)).}
\label{fig:further_experiments1}
\end{figure}

\subsubsection{Sample Size} Figure \ref{fig:further_experiments1} a) shows the effect of varying sample size on the performance of the offline RL algorithm TD3-BC. TD3-BC achieves comparable or better TBR and improvements to TIR for all sample sizes greater than or equal to 5e\textsuperscript{4} (approximately 100 days of glucose data). The poor TIR of TD3-BC for 1e\textsuperscript{4} samples of data is most likely due to the use of neural networks, as these algorithms perform most effectively with large quantities of data. For the greatest number of samples 5e\textsuperscript{5} ($>$ 1,000 days), TD3-BC achieves a \(2.2 \pm 1.1\%\) increase to TIR and a \(0.2 \pm 0.1\%\) reduction to TBR. The presented findings are consistent with the preliminary results in Fox, which show in a single adult patient that capable glucose control policies can be developed from as little as two months of data \cite{Fox2020MachineManagement}. This result is significant for the application of offline RL to future hybrid closed loop systems as 100 days represents a feasible timescale for data collection in patient populations. This sample represents less than one tenth of the data required for online RL approaches to surpass the PID controller in glucose control \cite{Fox2020DeepControlb,Zhu2021BasalValidation}. Greater sample efficiency could be achieved in future work by adopting a transfer learning approach, such as in Fox et al. \cite{Fox2020DeepControlb}. Under this method, general control strategies could be learnt by grouping patients by age or other demographic factors and training a general offline RL algorithm on this cohort. The pre-trained model could then be trained further on the target patient to achieve greater personalisation.   

\subsubsection{Suboptimal Demonstrations} Figure \ref{fig:further_experiments1} b) shows the dependency of TD3-BC performance on the quality of the training demonstrator. In all instances, TD3-BC can be seen to improve the control of the demonstrator by a margin of at least \(1.3 \pm 1.2\%\) to TIR. It also yielded a reduction of \(0.3 \pm 0.2\%\) to TBR when trained on the 10\textsuperscript{th} ranked PID policy. The most significant difference is observed for the 20\textsuperscript{th} ranked policy, whereby TD3-BC increases TIR by \(3.9 \pm 1.4\)\%, improves CV by \(3.4 \pm 1.1\%\) and reduces TBR by \(4.7 \pm 0.6\%\). An improvement of this magnitude could significantly improve the health outcomes of the user without the need for manually altering PID parameters \cite{Beck2019ValidationTrials}. The performance of the TD3-BC approach does decline by a significant margin of \(13.2 \pm 1.4\%\) to TIR and \(8.1 \pm 0.7\%\) for TBR between the 1\textsuperscript{st} and 20\textsuperscript{th} PID demonstrator. Therefore, achieving glycemic targets with offline RL in a hybrid closed loop system would still be largely dependent on the performance of the demonstrator in the training data. 

\begin{figure}[H]
\hspace{-7mm}
\centering
\includegraphics[width=0.85\textwidth]{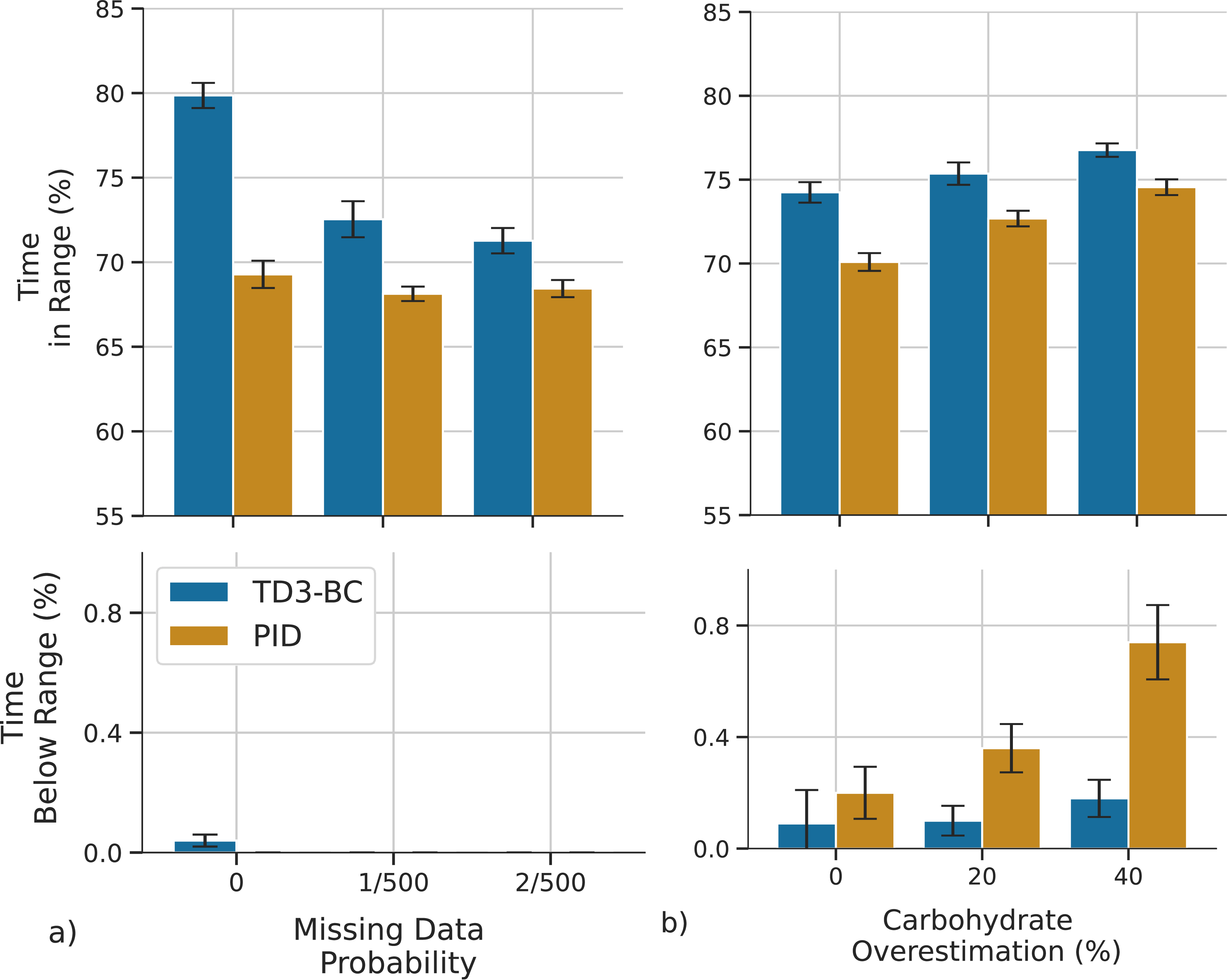}
\caption{ a) TD3-BC and PID performance with varying frequencies of missing blood glucose data. Time in the target glucose range (TIR) improvements are more modest with greater likelihood of missing data. b) TD3-BC and PID performance for varying levels of bolus overestimation. TD3-BC can be seen to avoid the pitfalls of the PID algorithm; experiencing no significant increase to time-below-range (TBR) for greater overestimation. Wilcoxon signed-rank tests confirmed the significance of the TD3-BC TIR improvement across all missing data likelihoods and bolus overestimation biases (\(p < 0.05\)).}
\label{fig:further_experiments2}
\end{figure}

\subsubsection{Missing Data}
Figure \ref{fig:further_experiments2} a) shows the effect of missing data on the effectiveness of TD3-BC. As before, TD3-BC yields a consistent improvement to TIR of at least \(2.83 \pm 0.9\%\), with no associated increase to TBR. However, the reduction in performance associated with the addition of missing data is particularly significant, resulting in a decrease of \(7.32 \pm 1.3\%\) to TIR. This may suggest that work is needed to improve the robustness of the TD3-BC algorithm to missing samples if real-world datasets are to be fully utilised. In this implementation, missing measurements were relabelled with the target blood glucose value to encourage the PID demonstrator to not take large insulin doses without accurate input data. However, in the state representation utilised by TD3-BC this replacement value is indistinguishable from a true blood glucose measurement and may have caused the performance degradation. In moving towards practical hybrid closed loop systems, it may be necessary to label these states more clearly or use an offline RL approach that is capable of incorporating missing values. 

\subsection{Safety Challenges of Offline Reinforcement Learning in Glucose Control}
\label{sec:safety_experiments}

PID was selected as the safest benchmark method as this algorithm has been extensively evaluated in real patients and has been approved for clinical use in hybrid closed loop systems across the world \cite{Weisman2017EffectTrials,Leelarathna2021Hybrid2021}. 

\subsubsection{Meal Overestimation} Figure \ref{fig:further_experiments2} b) compares the ability of PID and TD3-BC in correcting for overestimations in meal boluses. The TD3-BC approach can be seen to improve glucose control in both TIR and TBR across all levels of overestimation. This is particularly significant when considering a bolus overestimation of 40\%, which reduces TBR by \(0.6 \pm 0.2\%\). Mealtime miscalculation represents a frequent problem for people with T1D. A study on the accuracy of bolus calculations concluded that approximately 82\% of participants overestimated the carbohydrate content of common food choices, with the mean overestimation amount being 40\% \cite{Meade2016AccuracyAdults}. In an in vivo setting, this would allow for the correction of bias in bolus dosing without the inherent risk of using trial-and-error to alter mealtime calculations or PID parameters manually. Counter-intuitively, TIR increases across both algorithms for greater levels of carbohydrate overestimation. This increase is caused by higher levels of insulin-on-board in the patient, resulting in smaller post-meal blood glucose peaks, but also a greater susceptibility to low blood glucose events.    

\begin{figure}[H]
\hspace{-7mm}
\centering
\includegraphics[width=0.85\textwidth]{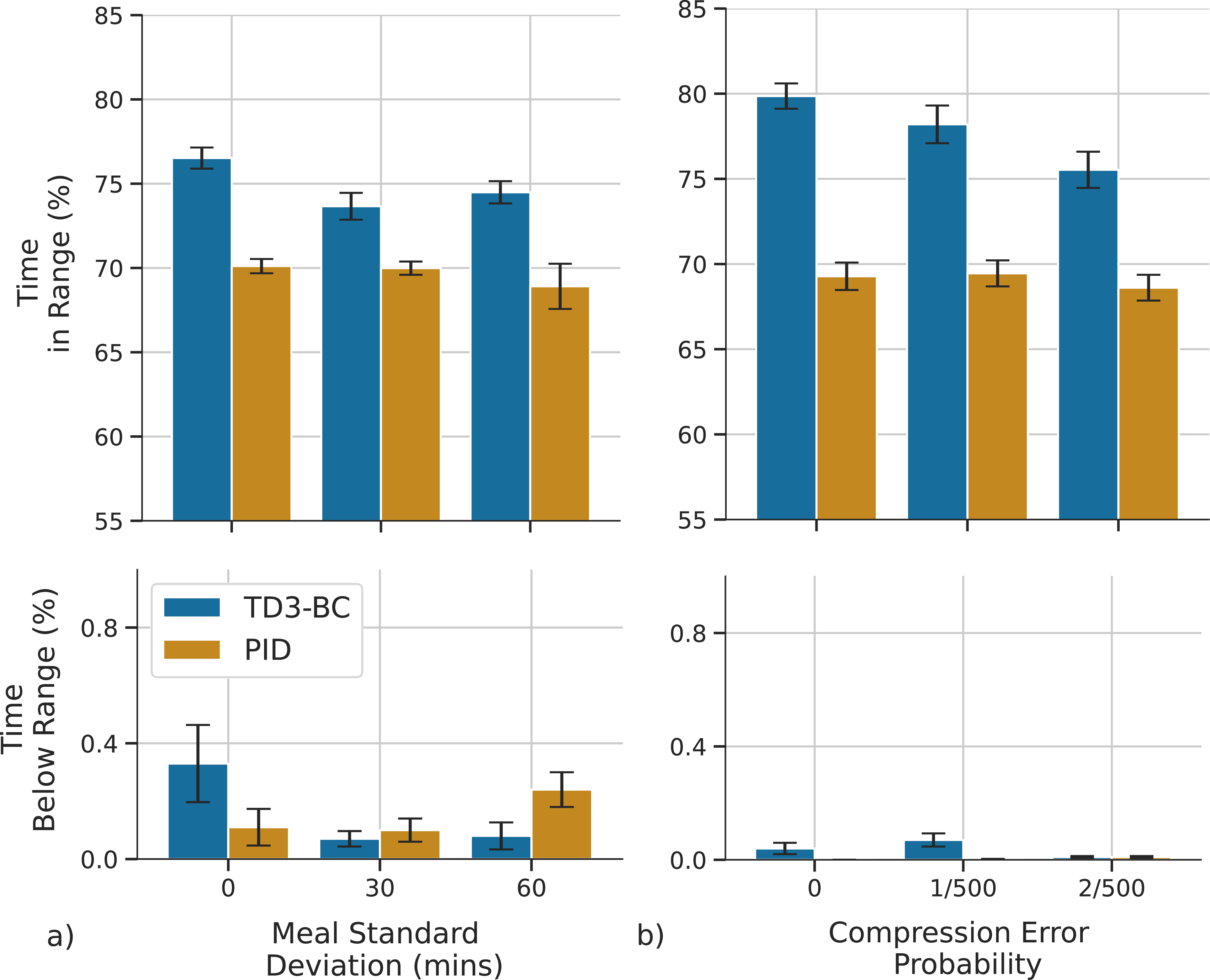}
\caption{a) TD3-BC and PID performance in relation to greater uncertainty in meal times, where meal standard deviation defines the observed deviation in meal time from the mean. TD3-BC achieves significantly better performance when standard deviation is 60 minutes; significantly increasing time-in-range (TIR) and reducing time-below-range (TBR) over the evaluation period. b) TD3-BC and PID performance with varying frequencies of compression low events. TD3-BC improves TIR significantly compared to the PID algorithm, however the margin reduces with greater frequency of compression events. Wilcoxon signed-rank tests confirmed the significance of the TD3-BC TIR improvement across all meal time uncertainties and compression low probabilities (\(p < 0.05\)).}
\label{fig:further_experiments3}
\end{figure}

\subsubsection{Irregular Meal Schedules} Figure \ref{fig:further_experiments3} a) examines the ability of TD3-BC to exploit regular meal schedules and adapt to greater uncertainty in meal timing. TD3-BC yields an improvement in TIR of at least \(3.7 \pm 1.2\%\) regardless of meal time standard deviation and without any significant worsening in TBR for non-zero meal deviation. The performance of TD3-BC evidently improves with the removal of meal uncertainty and snack events, as TIR is observed to increase by \(6.4 \pm 1.1\%\) (90 mins/day). This is also accompanied by an increase to TBR of \(0.2 \pm 0.2\%\), which may explain the observed improvement. This deterioration in policy could potentially be due to a lack of exploration in the training samples, resulting from the high meal regularity. When using real patient data, this flaw may become less apparent as this level of routine is unlikely to be achievable in a realistic patient setting. Adapting dosing policies to regular meal events may also transfer to other common routines in daily life such as work schedules or exercise plans provided there is sufficient contextual information in the state to intuit their occurrence. 

\subsubsection{Compression Error}
Figure \ref{fig:further_experiments3} b) assesses the robustness of TD3-BC to frequent compression errors in the CGM device. The TD3-BC approach yields an improvement of at least \(5.93 \pm 1.31\%\) regardless of event frequency. Compression lows are common occurrences at night time due to patients inadvertently applying pressure to their CGM sensors while sleeping. There is a strong association between poor nocturnal glycemic control and reduced sleep quality and duration \cite{Reutrakul2016SleepMeta-analysis}. This may suggest that utilising a more intelligent control algorithm, capable of responding more effectively to erroneous night time disturbances, may yield better sleep quality for patients. In this implementation, compression errors occurred randomly and were in no way linked to a patient's schedule. However, in a practical setting an offline RL algorithm may improve control further by identifying periods in which compression lows are more likely and using this information to more easily distinguish them from true changes in blood glucose.  

\section{Discussion and Conclusions}
\label{section:conclusions}

This work examined the application of offline RL for safer basal insulin dosing in hybrid closed loop systems. The experiments presented in this paper demonstrated that the offline RL approaches BCQ, CQL and TD3-BC were capable of learning effective control policies for adults, adolescents and children simulated within the UVA/Padova T1D model. In particular, the TD3-BC approach outperformed the widely-used and clinically validated PID algorithm across all patient age groups with respect to TIR, TBR and glycemic risk. The improvement was even more significant when TD3-BC was evaluated in potentially unsafe glucose control scenarios. Further experiments on TD3-BC also highlighted the ability of the approach to learn accurate and stable dosing policies from significantly smaller samples of patient data than those utilised in current online RL alternatives. 

This paper shows the potential of offline RL for creating safe and sample-efficient glucose control policies in people with T1D. In practice, the demonstrated offline RL method could be trained on an initial sample of patient data and then periodically retrained on data collected under the agent, allowing the algorithm to continually adapt to changes in the patient's insulin requirements. In moving towards an algorithm capable of full implementation within hybrid closed loop systems several avenues will first have to be explored. The most significant limitation of the presented evaluation is the use of the T1D simulator. As previously mentioned, these environments only capture a fraction of the complexity involved in realistic blood glucose dynamics; neglecting events such as stress, activity and illness. To confirm the scalability of offline RL approaches to more complex environments, algorithms will have to be trained and evaluated on real samples of retrospective patient data such as those available via the JCHR repository \cite{JaebCentreforHealthResearchJAEBDatasets}. This will require building on the current state-of-art in the offline evaluation of RL algorithms \cite{Fu2021BenchmarksEvaluationb}. This poses a particular challenge in glucose control, as the nature of the task means the effect of actions may only become apparent in some instances over several days, requiring blood glucose to be modelled over significantly longer prediction horizons than are currently deployed in commercial systems. 

In addition, further work will need to be done to provide safety assurances for deep learning driven hybrid closed loop systems. Although, the presented approach demonstrates significantly improved stability compared to prior online RL algorithms and the performance was validated on thousands of days of simulation, no guarantees can be made for the actions of the agent in a given scenario. This may make achieving regulatory approval challenging, especially as the agent takes actions on behalf of the patient rather than providing decision-support. In moving towards clinical usage, the presented algorithm would likely be most effective as one of many components in a hybrid closed loop device, with additional safeguarding systems in place for identifying harmful actions and providing reliable control policies.   
Future work could include validating the method on simulated populations with type 2 diabetes, building on offline RL methods to incorporate online learning for continuous adaption of control policies or incorporating features such as interpretability or integration of prior medical knowledge, which may ease the transition from simulation to clinical use.

\section{Conflict of Interest Statement}
None.

\section{Acknowledgements} 
This work was supported by the EPSRC Digital Health and Care Centre for Doctoral Training (CDT) at the University of Bristol (UKRI grant no. EP/S023704/1). 

% ----------------------------------------

%% The Appendices part is started with the command \appendix;
%% appendix sections are then done as normal sections
%% \appendix

%% \section{}
%% \label{}

%% If you have bibdatabase file and want bibtex to generate the
%% bibitems, please use
%%
\bibliographystyle{elsarticle-num} 
\bibliography{references.bib}

\end{document}